\documentclass[10pt,twocolumn,letterpaper]{article}

\usepackage{cvpr}
\usepackage{times}
\usepackage{epsfig}
\usepackage{graphicx}
\usepackage{amsmath}
\usepackage{amssymb}
\usepackage{booktabs}
\usepackage{multirow}
\usepackage{threeparttable}
\usepackage{bbm}
\usepackage{subfig}
\usepackage{algorithm}
\usepackage{algpseudocode}
\usepackage[pagebackref=true,breaklinks=true,letterpaper=true,colorlinks,bookmarks=false]{hyperref}

\usepackage{enumitem}
\usepackage{mmstyle}

\newcommand{\tmF}{\widetilde{\mF}}
\newcommand{\tvf}{\tilde{\vf}}

\cvprfinalcopy 

\def\cvprPaperID{250} 

\graphicspath{{figures/}}

\ifcvprfinal\fi
\begin{document}

\title{Unifying Identification and Context Learning for Person Recognition}

\author{
Qingqiu Huang, ~~ Yu Xiong, ~~ Dahua Lin\\
CUHK-SenseTime Joint Lab, The Chinese University of Hong Kong\\
{\tt\small \{hq016, xy017, dhlin\}@ie.cuhk.edu.hk}
}

\maketitle
\thispagestyle{empty}



\begin{abstract}
Despite the great success of face recognition techniques,
recognizing persons under unconstrained settings remains
challenging. Issues like profile views, unfavorable lighting,
and occlusions can cause substantial difficulties.
Previous works have attempted to tackle this problem by exploiting the context,
\eg~clothes and social relations.
While showing promising improvement, they are usually limited in two
important aspects,
relying on simple heuristics to combine different cues
and separating the construction of context from people identities.
In this work, we aim to move beyond such limitations and
propose a new framework to leverage context for person recognition.
In particular, we propose a Region Attention Network, which is learned
to adaptively combine visual cues with instance-dependent
weights.
We also develop a unified formulation, where the social contexts
are learned along with the reasoning of people identities.
These models substantially improve the robustness
when working with the complex contextual relations in
unconstrained environments.
On two large datasets, PIPA~\cite{zhang2015beyond} and Cast In Movies (CIM),
a new dataset proposed in this work, our method consistently
achieves state-of-the-art performance under multiple evaluation policies.
\end{abstract}


\section{Introduction}
\label{sec:introduction}

\begin{figure}[t]
	\centering
	\includegraphics[width=\linewidth]{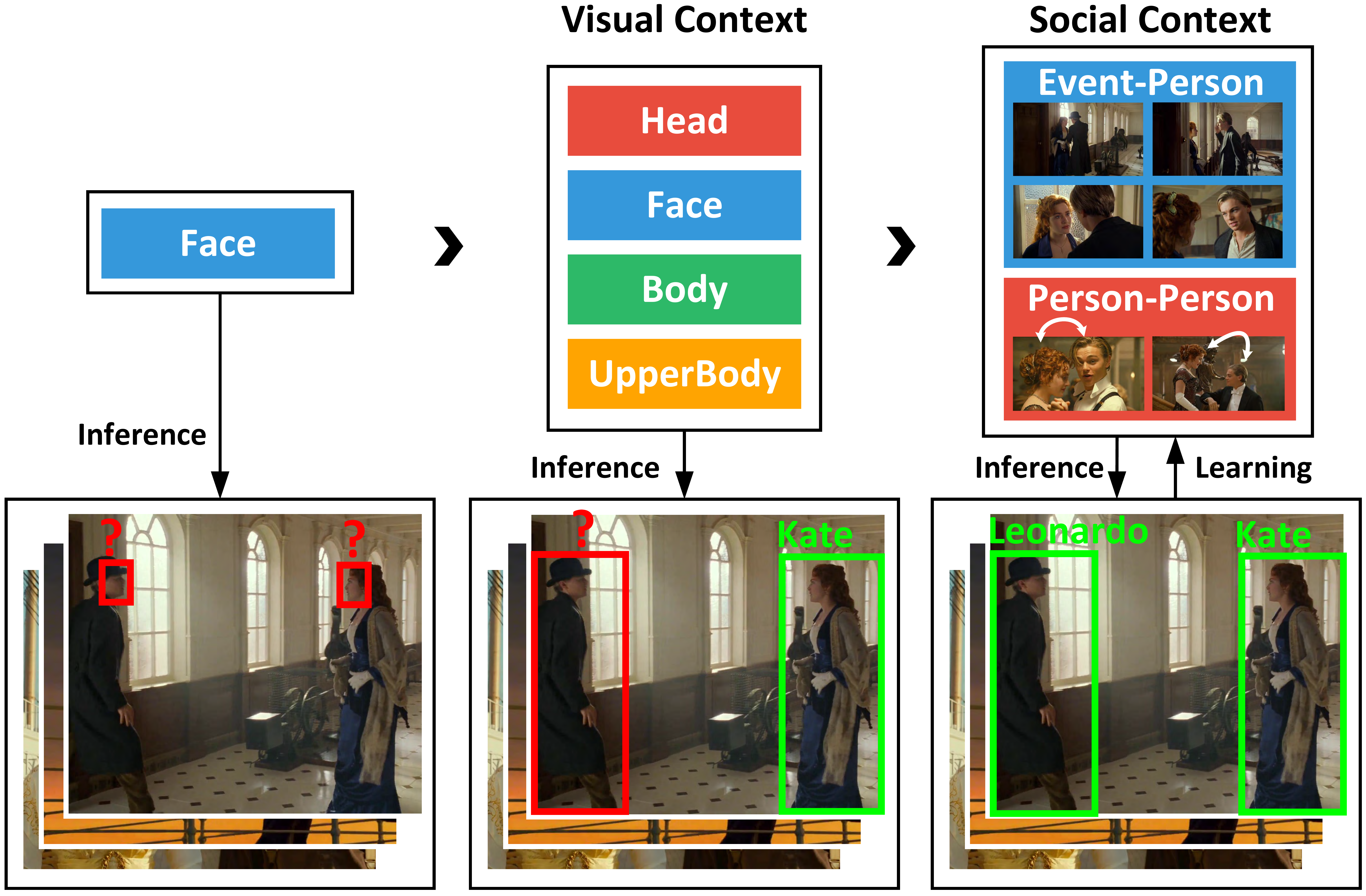}
	\caption{\small
		Person recognition under unconstrained settings remains a very challenging problem.
		Inference purely by face recognition techniques would fail in many cases.
		We propose a framework to tackle this problem,
		which combines \emph{visual context} with adaptive weights
		and unifies person recognition with \emph{social context} learning.
	}
	\vspace{-10pt}
	\label{fig:teaser}
\end{figure}


Person recognition is a key task in computer vision and has been
extensively studied over the past decades.
Thanks to the advances in deep learning, recent years have witnessed
remarkable progress in face recognition techniques~\cite{taigman2014deepface,schroff2015facenet,sun2015deepid3,liu2015targeting}.
On LFW~\cite{huang2007labeled}, a challenging public benchmark,
the accuracy has been pushed to over $99.8\%$~\cite{liu2015targeting}.
Nonetheless, the success on benchmarks does not mean that
the problem has been well solved.
Recent studies~\cite{zhang2015beyond,joon2015person,li2016multi,li2016sequential}
have shown that recognizing persons under an unconstrained setting
remains very challenging.
Substantial difficulties arise in unfavorable conditions,
\eg~when the faces are in a non-frontal position,
subject to extreme lighting, or too far away from the camera.
Such conditions are very common in practice.


The difficulties above are essentially due to the fact that
facial appearance is highly sensitive to environmental conditions.
To tackle this problem, a natural idea is to leverage another
important source of information, namely the \emph{context}.
It is our common experience that we can easily recognize
a familiar person by looking at the wearing, the surrounding environment,
or the people who are nearby.
On the other hand, cognitive neuroscience studies~\cite{chun1998contextual,adolphs2003cognitive,labar2006cognitive}
have shown that \emph{context} plays a crucial role when we,
as human beings, recognize a person or an object.
A familiar context often allows much greater accuracy in recognition.


Exploiting context to help recognition is not a new story in computer vision.
Previous efforts mainly follow two lines.
The first line of research~\cite{anguelov2007contextual,gallagher2008clothing,zhang2015beyond,joon2015person} attempts
to incorporate additional visual cues, \eg~clothes and hairstyles,
as additional channels of features.
The other line, instead, focuses on social relationships,
\eg~group priors~\cite{davis2005towards,song2006context} or people co-occurrence~\cite{brenner2014joint,wang2010seeing}.
There are also studies that try to integrate both visual cues
and social relations~\cite{li2016sequential,li2016multi}.

Whereas previous works have shown the utility of context in
person recognition, especially in unconstrained environments,
a key question remains open, that is,
\emph{how to discover and leverage contexts robustly}.
Specifically, existing methods usually rely on simple heuristics,
\eg~feature-based clustering, to establish contextual priors,
and hand-crafted rules to combine contextual cues from different
channels.
Moreover, the construction of the context model is typically done
\emph{separately} and \emph{before} person identification.
The limitations of such approaches lie in two important aspects:
(1) Heuristics designed manually are difficult to capture the diversity
and complexity in unconstrained context modeling.
(2) The identities of the people in a scene are also an important part
of the context. Constructing the context separately would lose
this significant connection.


In this work, we aim to explore more effective ways to leverage the context
in person recognition (see Figure~\ref{fig:teaser}). Inspired by previous efforts, we consider two kinds of
contexts, namely the visual context, \eg~additional visual cues,
and the social context, \eg~the events that a person often attends.
But, we move beyond the limitations of existing methods, by considering
context learning and person identification as
a unified process and solving both jointly.
Driven by this idea, we propose novel methods for leveraging visual
and social contexts respectively.
Particularly, we develop a \emph{Region Attention Network}, which is
\emph{learned end-to-end} to combine various visual cues adaptively
with instance-dependent weights.
We also develop a unified formulation, where the social context model
is learned online \emph{jointly} with the reasoning of people identities.
As a by-product, the solution to this problem also comes with
a set of \emph{``events''} discovered from the given photo collection
-- an event not only share similar scenes but also
a consistent set of attendants.


On PIPA~\cite{zhang2015beyond}, a large public benchmark for person recognition,
our proposed method consistently outperform existing methods, under
all evaluation policies. Particularly, in the most challenging
\emph{day split}, our method raised the state-of-the-art performance
from $59.77\%$ to $67.16\%$.
To assess our method in more diverse settings and to promote future research
on this topic, we construct another large dataset,
\emph{Cast In Movies (CIM)}, by annotating the characters in $192$ movies.
This dataset contains more than $150K$ person instances and
$1218$ labeled identities.
Our approach also demonstrated its effectiveness on \emph{CIM}.


Our contributions mainly lie in three aspects:
(1) For visual context, we propose a \emph{Region Attention Network},
which combines visual cues with instance-dependent weights.
(2) For social context, We propose a \emph{unified formulation} that couples
context learning with people identification.
It also discovers events from photo collections automatically.
These two techniques together result in remarkable performance gains
over the state-of-the-art.
(3) We construct \emph{Cast In Movies (CIM)},
a large and diverse dataset for person recognition research.


\section{Related Work}
\label{sec:related}

\vspace{-4pt}
\paragraph{Early efforts.}
The significance of context in person recognition has long been recognized
by the vision community.
Early methods mainly tried to use additional visual cues,
such as clothing~\cite{anguelov2007contextual,gallagher2008clothing},
or additional metadata~\cite{davis2005towards}.
Yet, the improvement was limited.
Later, more sophisticated
frameworks~\cite{song2006context,anguelov2007contextual,lin2010joint}
that integrate multiple cues (clothing, timestamps, scenes, etc)
have been developed.
Some of these works~\cite{anguelov2007contextual,lin2010joint} formulated
the task as a joint inference process over a Markov random field
and obtained further performance gains.
Note that these MRF-based methods assume the same set of people and social
relations in both training and testing, and thus the learned models are
difficult to generalize to new collections.

\vspace{-12pt}
\paragraph{Recent efforts.}
The rise of deep learning has led to new innovations on this topic.
\emph{Zhang et al.}~\cite{zhang2015beyond} proposed a Pose Invariant
Person Recognition method (PIPER), which combines three types of
visual recognizers based on ConvNets,
respectively on face, full body, and poselet-level cues.
The PIPA dataset published in~\cite{zhang2015beyond} has been widely
adopted as a standard benchmark to evaluate person recognition methods.
\emph{Oh et al.}~\cite{joon2015person} evaluated the effectiveness of
different body regions, and used a weighted combination of the scores
obtained from different regions for recognition.

Recently, \emph{Li et al.}~\cite{li2016multi} proposed a multi-level
contextual model, which integrates person-level, photo-level and group-level contexts.
Although this framework also considers the combination of visual cues
and social context, it differs from ours essentially in two key aspects:
(1) The visual cues are combined with a simple heuristic rule, instead
of a learned network.
(2) The groups are identified by spectral clustering of scene features
\emph{before} person recognition, as a \emph{separate} step.
Our framework, instead, formulates event discovery and people identification
as a unified optimization problem and solves both jointly.

Another way of integrating both visual cues and social relations was
proposed in~\cite{li2016sequential}. This work formulates the recognition
of multiple people into a sequence prediction problem and tries to capture
the relational cues with a recurrent network. As there is no inherent order
among the people in a scene, it is unclear how a sequential model can
capture their relations.
Note that we compared the proposed method with all of the four methods above in
our experiments on PIPA. As we shall see in Section~\ref{sec:experiment},
our method consistently outperforms them under all evaluation policies.


\vspace{-12pt}
\paragraph{Person Re-identification.}
Another relevant task is
person re-identification~\cite{prosser2010person,zhao2013unsupervised,li2014deepreid},
which is to match pedestrian images from different cameras, within a relatively
short period.
This task is essentially different, where visual cues
are likely to remain consistent and social context is weak.
General person recognition, instead, requires recognizing
across events, where visual cues may vary significantly
and thus the social context can be crucial.


\begin{figure*}[thb]
	\centering
	\includegraphics[width=0.9\linewidth]{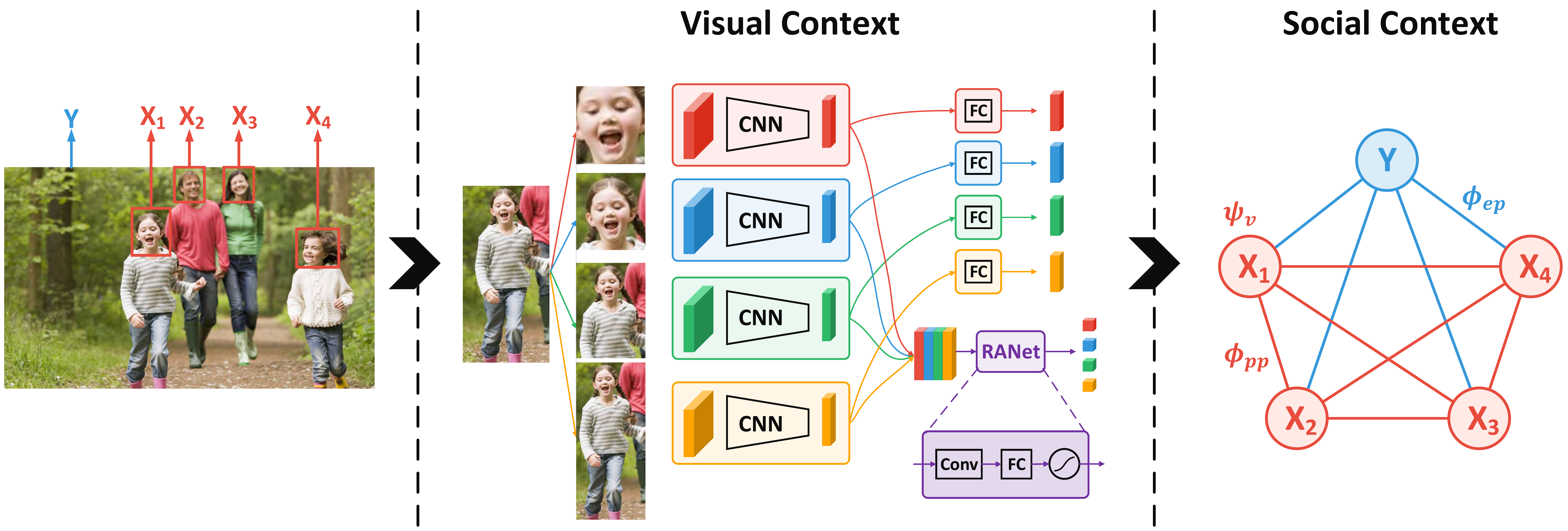}
	\vspace{-10pt}
	\caption{\small
		Our whole framework. We propose a Region Attention Network to get instance-dependent
		weights for visual context fusion 
		and develop a unified formulation that join social context learning,
		including event-person relations and person-person relations, with person recognition.
    }
    \label{fig:framework}
    \vspace{-10pt}
\end{figure*}

\section{Methodology}
\label{sec:method}

In general, the task of person recognition in a photo collection can be
formalized as follows.
Consider a collection of photos $I_1, \ldots, I_M$,
where $I_m$ contains $N_m$ \emph{person instances}.
All person instances are divided into two disjoint subsets,
the \emph{gallery set}, in which the instances are all labeled
(\ie~their identities are provided).
and the \emph{query set}, in which the instances are unlabeled.
The task is to predict the identities for those instances in the
query set.

As discussed, person recognition in an unconstrained setting is
very challenging. In this work, we leverage two kinds of contexts,
the \emph{visual context} and the \emph{social context}.
Particularly, the \emph{visual context} involves different regions
of the person instances, including
\emph{face}, \emph{head}, \emph{upper body}, and \emph{whole body}.
These regions often convey complementary cues for visual matching.
The \emph{social context}, instead, captures the social behavior
of people, \eg~the events they usually attend or the people whom
they often stay with.
It is worth noting that unlike visual cues,
the social relations are reflected collectively by multiple photos and
can not be reliably derived from a single photo in isolation.

\subsection{Framework Overview}

We devise a framework that incorporates both the visual context
and the social context for person recognition.
As shown in Figure~\ref{fig:framework}, the framework recognizes
the identities for all instances in the query set \emph{jointly},
in two stages.
\begin{enumerate}[leftmargin=*,labelindent=0pt,itemsep=0pt]
\item \textbf{Visual matching.} This stage computes
a \emph{matching score} for each pair of instances.
For this, a \emph{Region Attention Network} is learned to adaptively
combine the visual cues from different regions,
with instance-dependent weights.

\item \textbf{Joint optimization.}
This is the key stage of our framework.
In this stage, the \emph{social context model},
which captures both \emph{event-people} and \emph{people-people} relations,
will be jointly learned along with the identification of query instances,
by solving a unified optimization problem.
\end{enumerate}

\subsection{Visual Matching}
\label{sec:vmatch}

We combine the visual observations from different regions
to compute the \emph{matching score} between two instances.
Particularly, we consider four regions:
\emph{face}, \emph{head}, \emph{upper body}, and \emph{whole body}.
These regions are often complementary to each other.
This strategy has also been shown to be effective in previous work~\cite{zhang2015beyond,joon2015person,li2016multi,li2016sequential}.

However, existing methods mostly adopt uniform weighting schemes,
where each region is assigned a fixed weight that is shared by all instances.
Let $s(i, j)$ be the overall matching score between instances $i$ and $j$,
and $s^r(i, j)$ be the matching score based on the $r$-th region.
Then, such a scheme can generally be expressed as
\begin{equation}
    s(i, j) = \sum_{r=1}^R w^r s^r(i, j),
\end{equation}
where $R$ is the number of distinct regions.
The weights $\{w^r\}$ are often decided by empirical rules~\cite{li2016multi}
or optimized over a validation set~\cite{zhang2015beyond,joon2015person,li2016sequential}.

\begin{figure}[t]
	\centering
	\subfloat[\label{fig:visual_case_1}]{\includegraphics[width=0.48\linewidth]{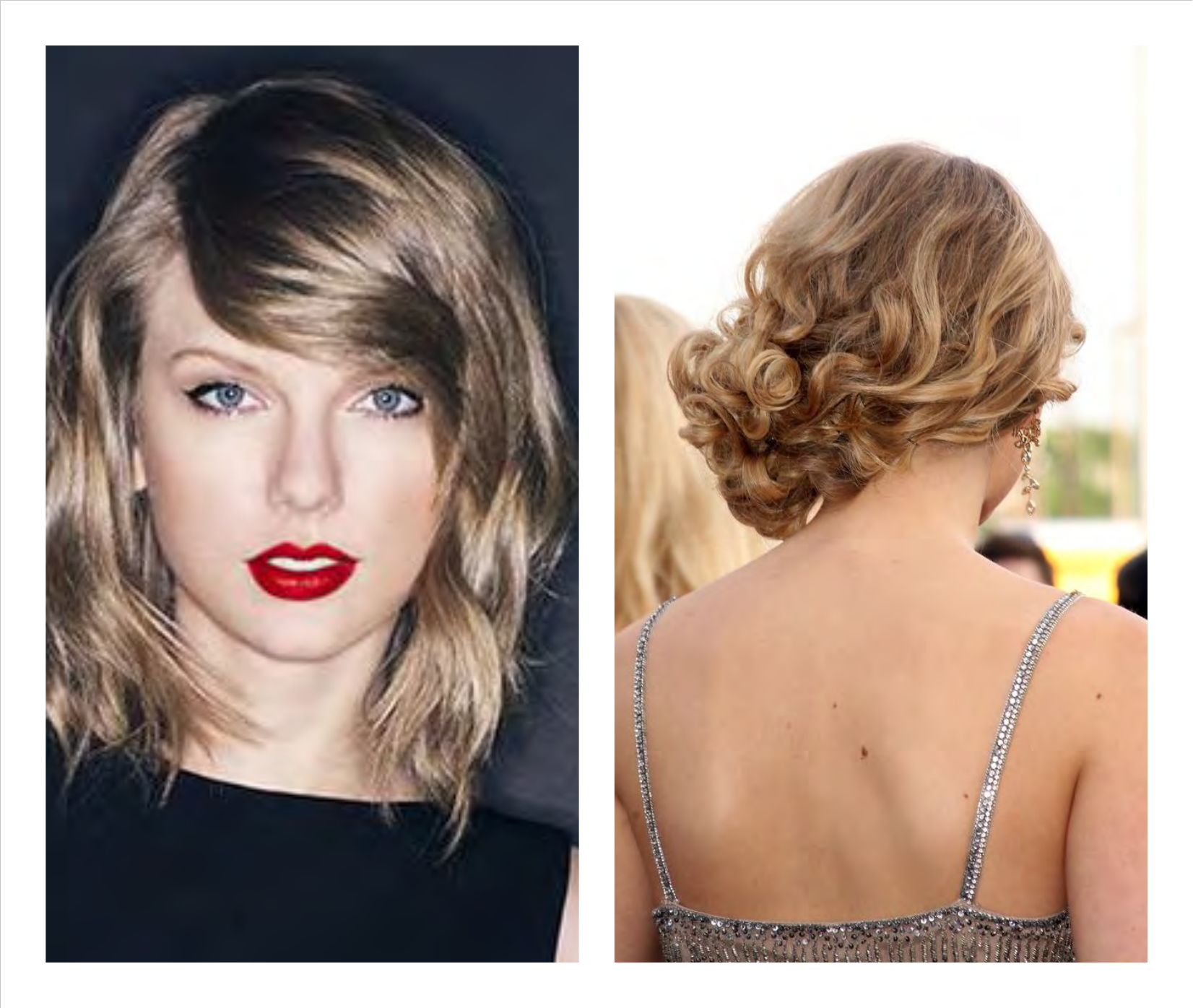}} \hfill
	\subfloat[\label{fig:visual_case_2}]{\includegraphics[width=0.48\linewidth]{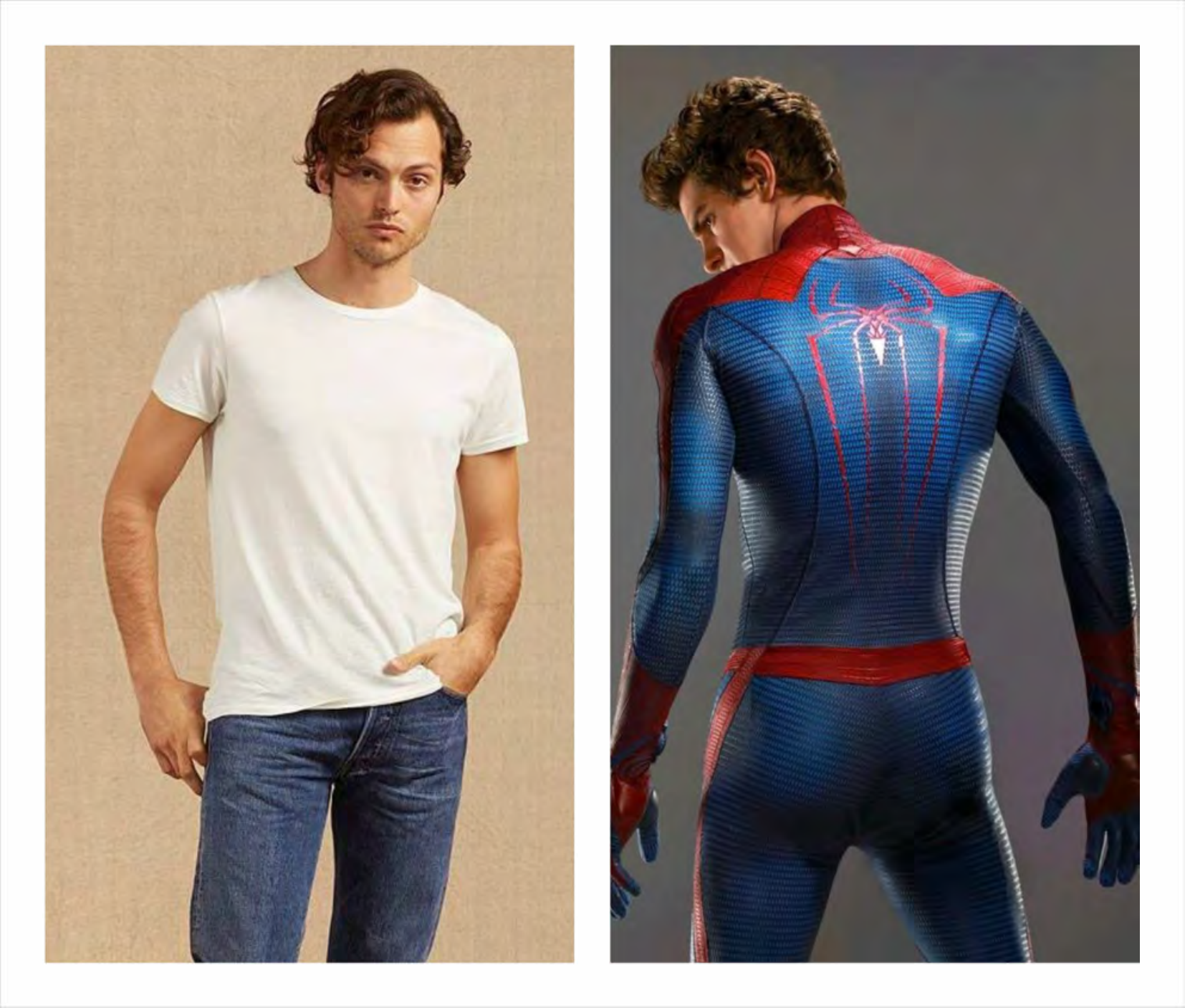}}
	\vspace{-10pt}
	\caption{\small
		Here are examples to show the necessity of instance-dependent weights for
		recognition.
		(a) shows that some of the regions may be out of scope,
		like ``body'' of the first instance and ``face'' of the second instance.
		(b) shows that the contributions of parts vary across instances.
		The contribution of ``face'' for the first man is obviously
		more significant than that for the second one.
	}
	\vspace{-10pt}
	\label{fig:visual_case}
\end{figure}

The uniform schemes as described above are limited in two aspects,
as illustrated in Figure~\ref{fig:visual_case}.
(1) Some regions may be invisible for an instance.
The missing of such regions may be due to various reasons,
\eg~limited scope of the camera and occlusion.
With a uniform scheme,
one would be forced to locate the missing parts with rigid rules and
compute matching scores for them, which
often leads to inaccurate results.
(2) The contributions of different parts vary significantly
across instances. For example, the facial features play a key role
when the frontal face is visible. However, when we can only see one's
back, we will have to resort to the clothing in the body region.
A uniform scheme can not effectively handle such variations.

We propose to tackle this problem using \emph{instance-dependent} weights,
where the weight of a region is determined by whether it is visible
and how much it contributes.
Specifically, given an instance,
we get the bounding boxes of the regions by either the annotation of the dataset or
detectors,
then resize each region to a standard size, and apply region-specific CNNs to extract
their features.

To combine these features adaptively,
we devise a \emph{Region Attention Network (RANet)} as shown
in Figure~\ref{fig:framework} to compute the fusion weights.
Here, the RANet is a small neural network that takes
the stacked features from all regions as input, feeds them
through a convolution layer, a fully-connected layer, and a sigmoid layer,
and finally yields four positive coefficients as the region weights.
Then the combined \emph{matching score} is given by
\begin{equation} \label{eq:vmatch}
    s(i, j) = \sum_{r=1}^R w_i^r w_j^r s^r(i, j).
\end{equation}
Here,
$w_i^r$ and $w_j^r$ are instance-dependent weights of the $r$-th region
respectively for instances $i$ and $j$;
$s^r(i, j)$ is the cosine similarity between the corresponding features.
We use the product $w_i^r w_j^r$ to weight a region score, which reflects
the rationale that a region type should be active
only when it is clearly visible in both instances.
All region-specific CNNs together with the RANet are jointly trained
in an \emph{end-to-end manner} with the cross-entropy loss.

\begin{figure}[t]
	\centering
	\subfloat[\label{fig:social_case_1}]{\includegraphics[width=0.48\linewidth]{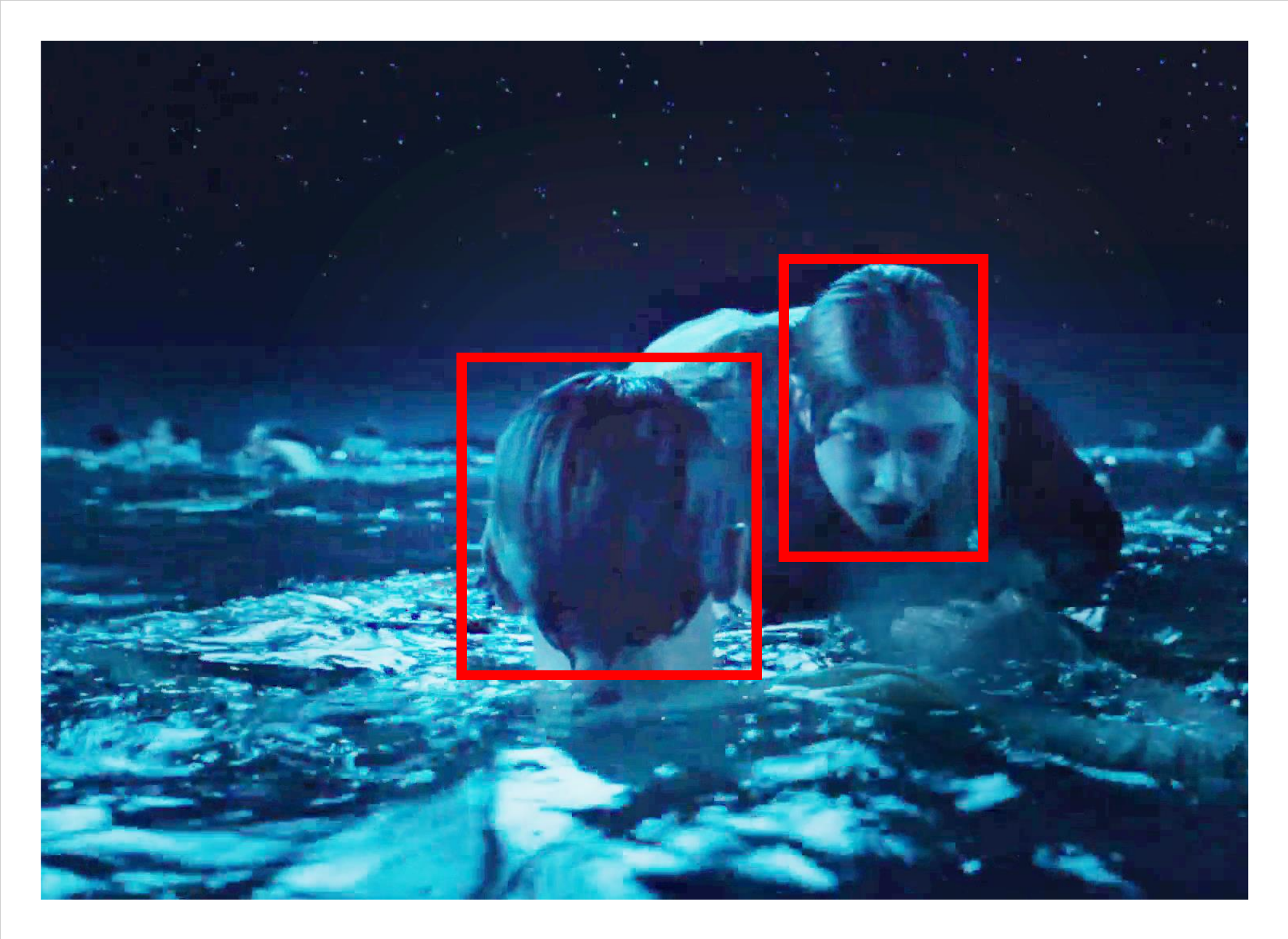}} \hfill
	\subfloat[\label{fig:social_case_2}]{\includegraphics[width=0.48\linewidth]{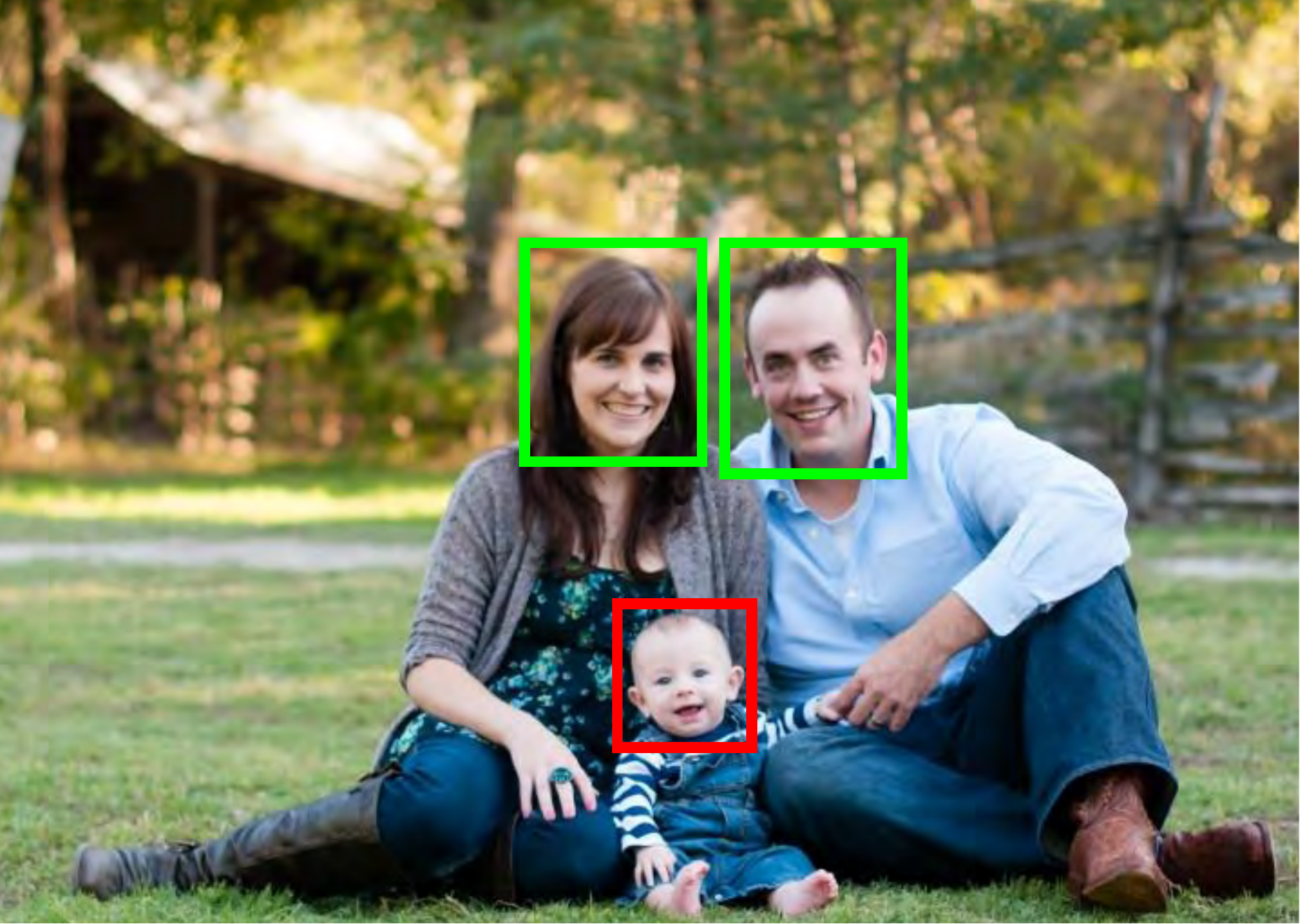}}
	\vspace{-10pt}
	\caption{\small
		Some instances (with red boxes) are very difficult to recognize purely by their
		appearance. But we can leverage the social context to help.
		(a) If we know that the photo belongs to the event -- ``People cry for help after
		Titanic sunk'', then the probability to recognize them as \emph{Leonardo} and \emph{Kate} will become higher.
		(b) Sometimes it's easy for us to get other instances' identities (with green boxes) in the same photo.
		So we can infer the red ones' identities by the person-person relation.
	}
	\vspace{-10pt}
	\label{fig:social_case}
\end{figure}

\subsection{Unified Formulation with Social Context}

In an unconstrained environment, certain instances are very difficult
to recognize purely based on their appearance. For such cases,
one can leverage the social context to help.
Specifically, the \emph{social context} refers to a set of social relations.
We consider two types of social relations:
\begin{enumerate}[leftmargin=*,labelindent=0pt,itemsep=0pt]
\item \textbf{Event-person relations.}
Generally, an \emph{event} can be conceptually understood as
an activity that occurs at a certain place with a certain set of attendants~\cite{xiong2015recognize}.
Over a large photo collection, an event may involve just a small fraction
of the people. Hence, an event can provide a strong prior for
recognition if we can infer the event that a photo is capturing,
as illustrated by the photo in Figure~\ref{fig:social_case_1}.

\item \textbf{Person-person relations.}
It is commonly observed that certain groups of people often stay together.
For such groups, the presence of a person may indicate the presence
of others in the group, as illustrated by the photo
in Figure~\ref{fig:social_case_2}.
Note that \emph{person-person relations} are complementary to
the \emph{event-person relations}, as such relations do not
require the match of scene features.
\end{enumerate}

\vspace{-10pt}
\paragraph{Unified Objective.}
Taking both the visual context and the social context into account,
we can formulate a unified optimization problem where
person identifications are coupled with event association and contextual relation learning.
The objective function of this problem is given below:
\begin{multline} \label{eq:jointobj}
    J(\mX, \mY; \tmF, \mP, \mQ \mid \mS, \mF)
    = \psi_v(\mX | \mS) \\
    + \alpha \cdot \phi_{ep}(\mY, \mX; \tmF, \mP | \mF)
    + \beta \cdot \phi_{pp}(\mX; \mQ).
\end{multline}
The notations involved here are described below:
\begin{enumerate}[leftmargin=*,labelindent=0pt,itemsep=0pt]
\item $\mX \in \Rbb^{L \times N}$ captures all people identities,
where $L$ is the number of distinct identities and
$N$ is the number of person instances.
Each column of $\mX$, denoted by $\vx_j$, is the identity indicator
vector for the $j$-th instance.
\item $\mY \in \Rbb^{K \times M}$ is the association matrix between
photos and events, where $K$ is the number of events and $M$
is the number of photos.
The $i$-th column of $\mY$, denoted by $\vy_i$, is the event indicator
vector for the $i$-th photo.
In particular, $y_i^k \triangleq \mY(i, k) \in \{0, 1\}$ indicates
whether the $i$-th photo is from the $k$-th event.
\item $\mS \in \Rbb^{N \times N}$ denotes the matrix of
pairwise visual matching scores, derived by Eq.\eqref{eq:vmatch}.
\item $\mF \in \Rbb^{D_f \times M}$ comprises the scene features
of all photos, where the $i$-th column $\vf_i$ is
a $D_f$-dimensional feature for the $i$-th photo. In this work,
we obtained $\vf_i$ for each photo with a CNN pretrained
on Places~\cite{zhou2014learning}.
\item $\tmF \in \Rbb^{D_f \times K}$
and $\mP \in \Rbb^{L \times K}$ are the parameters associated with
the \emph{events}. In particular,
the $k$-th column of $\tmF$, denoted by $\tvf_k$, is the prototype scene
feature for the $k$-th event;
and the $k$-th column of $\mP$, denoted by $\vp_k$, is a probability vector
that captures the person identity distribution of the $k$-th event.
\item $\mQ \in \Rbb^{L \times L}$ is a matrix that captures
the person-person relations. High value of $\mQ(l, l')$ indicates
that the identities $l$ and $l'$ are likely to co-occur in the same photo.
\end{enumerate}
Among these quantities,
the matching scores $\mS$ and the scene features $\mF$ are
provided in the visual analysis stage, while others are jointly
solved by optimizing this problem.

\vspace{-9pt}
\paragraph{Potential Terms.}
The joint objective in Eq.\eqref{eq:jointobj} comprises
three potential terms, which are introduced below.

\begin{enumerate}[leftmargin=*,labelindent=0pt,itemsep=0pt]
\item \textbf{Visual consistency:} $\psi_v(\mX | \mS)$ encourages
the consistency between person identities and the visual matching scores,
and is formulated as:
{\small
\begin{equation}
	\psi_v(\mX | \mS) =
	\sum_{j=1}^N \mathbf{s}_j^T \vx_j, \
	\text{ with }
	\mathbf{s}_j(l) = \max_{j' \in \cG_l} s(j, j'),
\end{equation}
}
where $\mathbf{s}_j \in \Rbb^L$,
$\cG_l$ refers to the set of gallery instances with label $l$,
and thus $\mathbf{s}_j(l)$ is the maximum matching score
of the $l$-th instance to those in $\cG_l$.

\item \textbf{Event consistency:}
$\phi_{ep}(\mY, \mX; \tmF, \mP | \mF)$ concerns about the assignments
of photos to events, and encourages them to be consistent in both
scenes and attendants.
This term is formulated as:
\begin{align} \label{eq:ep_term}
	\phi_{ep}(\mY, \mX; \tmF, \mP | \mF)
	= \sum_{i=1}^M \sum_{k=1}^K a_i^k y_i^k \notag \\
	\text{ with } a_i^k =
	\sum_{j \in \cI_i} \log(\vp_k)^T \vx_j - \|\vf_i - \tvf_k\|^2,
\end{align}
where, $\cI_i$ is the set of instance indexes for the $i$-th photo.
For each assignment of the $k$-th event to the $i$-th photo,
this formula evaluates (a) whether the people in the photo are frequent
attendants of the event (by $\vp_k^T \vx_j$) and (b)
whether the photo's scene feature match the event's scene prototype
(by $-\|\vf_i - \tvf_k\|^2$).

\item \textbf{People cooccurrence:}
$\phi_{pp}(\mX; \mQ)$ takes into account the person-person relations,
\ie~which identities tend to coexist in a photo. This term is
formulated as:
\begin{equation} \label{eq:pp_term}
	\phi_{pp}(\mX; \mQ) =
	\sum_{i=1}^M \sum_{j \in \cI_i} \sum_{j' \in \cI_i: j' \ne j}
	\vx_j^T \mQ \vx_{j'}.
\end{equation}
This formula considers all pairs of distinct instances in each image,
and sums up their person-person relation value.
In particular, if $\vx_j$ indicates label $l$ and $\vx_{j'}$ indicates
label $l'$, then $\vx_j^T \mQ \vx_{j'} = Q(l, l')$.
\end{enumerate}
To balance the contributions of these potential terms, we introduce
two coefficients $\alpha$ and $\beta$, which are decided via cross validation.

\subsection{Joint Estimation and Inference}
We solve this problem using coordinate ascent. Specifically, our algorithm
alternates between the updates of
(1) people identities ($\mX$),
(2) assignments of events to photos ($\mY$), and
(3) social relation parameters ($\tmF$, $\mP$, and $\mQ$).
These steps are presented below.

\vspace{-7pt}
\paragraph{Person Identification.}
Given both the event assignments $\mX$ and the social context parameters,
the inference of people identities can be done separately for each photo,
by maximizing the sub-objective as:
\begin{equation}
\label{eq:p_infer}
	  \sum_{j \in \cI_i} \mathbf{s}_j^T \vx_j
	+ \alpha \sum_{j \in \cI_i} \log(\vp_{\hat{y}_i})^T \vx_j
	+ \beta \sum_{j \in \cI_i} \sum_{j' \in \cI_i: j' \ne j} \vx_j^T \mQ \vx_j,
\end{equation}
where $\hat{y}_i$ indicates the assigned event.
Note that $\vx_j$ here is constrained to be an indicator vector, \ie~only one
of its entry is set to one, while others are zeros.
When there is only one person instance, its identity can be readily derived as
\begin{equation}
	\hat{x}_j = \argmax_{l} \ \mathbf{s}_j(l) + \alpha \vp_{\hat{y}_i}(l).
\end{equation}
When there are two or more instances,
we treat it as an MRF over their identities and solve them \emph{jointly}
using the max-product algorithm.

\vspace{-7pt}
\paragraph{Event Assignment.}
We found that
the granularity of the events has significant impact on the identification
performance. If we group the photos into coarse-grained events such that
each event may contain many people or scenes, then the event-person relations
may not be able to provide a strong prior.
However, for fine-grained events, it would be difficult to estimate
their parameters reliably. Hence, it is advisable to seek a good balance.

In this work, we use two parameters $\nu_{min}$ and $\nu_{max}$ to control
the granularity, and require that the number of photos assigned to
an event fall in the range $[\nu_{min}, \nu_{max}]$.
Then the problem of event assignment can be written as
\begin{align}
\max ~~~~ & \sum_{i=1}^M \sum_{k=1}^K  a_i^k y_i^k, \label{eq:eobj} \\
s.t. ~~~~ & \sum_{k=1}^K y_i^k \le 1 , ~~ \forall i = 1, \ldots, M, \label{eq:e_c1}\\
& \nu_{min} \le \sum_{i=1}^M y_i^k \le \nu_{max} , ~~ \forall k = 1, \ldots, K, \label{eq:e_c2}.
\end{align}
Here, $a_i^k$ Eq.\eqref{eq:eobj} follows Eq.\eqref{eq:ep_term}.
Eq.\eqref{eq:e_c1} enforces the constraint that each photo is associated to
at most one event;
Eq.\eqref{eq:e_c2} enforces the granularity constraint above.
This is a linear programming problem, and can be readily solved by an LP solver.
Also, the optima is guaranteed to be integral.

\vspace{-7pt}
\paragraph{Context Learning.}
As mentioned, the social context model,
which is associated with three parameters $\tmF$, $\mP$, and $\mQ$, are
learned along the inference of people identities and event assignments.
Given $\mX$ and $\mY$, we can easily derive the optimal
solution of the parameters listed above.

Specifically, for the scene prototypes in $\tmF$, we have the optimal
$\tvf_k$ (the $k$-th column of $\tmF$) given by
\begin{equation}
	\tvf_k
	= \argmin_{\vf} \sum_{i \in \cE_k} \|\vf_i - \vf\|^2
	= \frac{1}{|\cE_k|} \sum_{i \in \cE_k} \vf_i,
\end{equation}
where $\cE_k = \{i \mid a_i^k = 1\}$ refers to the set of photos
that are assigned to the $k$-th event.
For the identity distributions $\mP$, we have the optimal $\vp_k$
given (the $k$-th column of $\mP$) given by
\begin{equation}
	\vp_k
	= \argmax_{\vp \in \Sbb^L} \sum_{i \in \cE_k} \sum_{j \in \cI_i}
	\log(\vp)^T \vx_j,
\end{equation}
where $\Sbb$ is the $L$-dimensional probability simplex and $\vp \in \Sbb^L$
enforces that $\vp$ be a probability vector.
This is a maximum likelihood estimation over
all people who attend the $k$-th event, and its optimal solution is
\begin{equation}
	\vp_k = \left(\sum_{i \in \cE_k} |\cI_i| \right)^{-1}
	\sum_{i \in \cE_k} \sum_{j \in \cI_i} \vx_j.
\end{equation}
With both $\tvf_k$ and $\vp_k$, we can characterize an event with
both scene features and attendants.
For person-person relations,
the optimal $\mQ$ can be obtained by maximizing Eq.\eqref{eq:pp_term}
with $\mX$. Here, we enforce a constraint that $\mQ$ is normalized,
\ie~$\|\mQ\|_F = 1$. Then, the optimal solution is
\begin{equation}
	\mQ = \mQ' / \|\mQ'\|_F, \text{ with }
	\mQ' = \sum_{i=1}^M \sum_{j \in \cI_i} \sum_{j' \in \cI_i \backslash j}
	\vx_j \vx_{j'}^T.
\end{equation}
It is worth emphasizing that all sub-tasks presented above are
steps in the coordinate ascent procedure to optimize the unified objective
in Eq.\eqref{eq:jointobj}. We run these steps iteratively, and it usually
takes about $5$ iterations to converge.


\section{New Dataset: Cast In Movies}
\label{sec:dataset}

\begin{figure}[!ht]
	\centering
	\includegraphics[width=\linewidth]{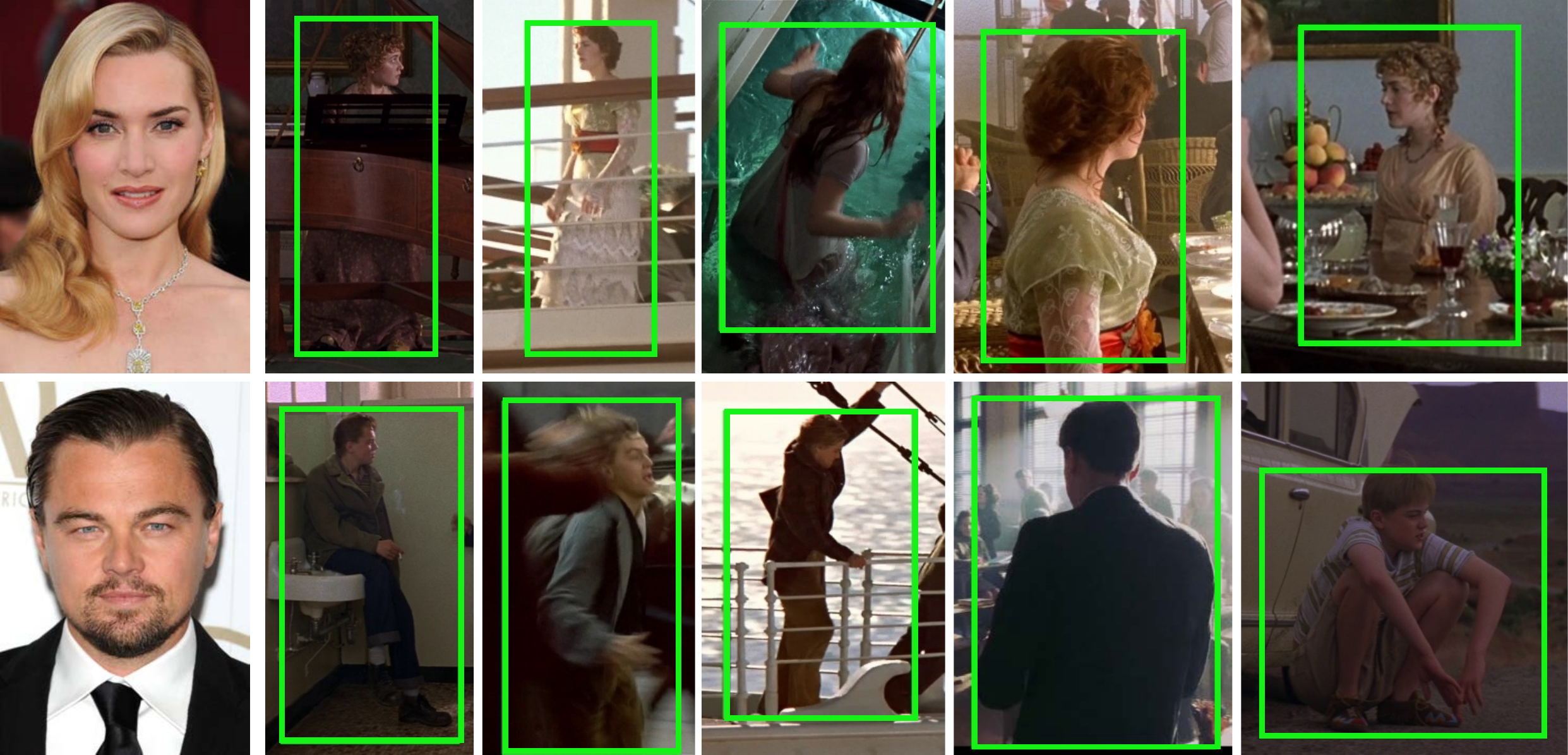}
	\vspace{-15pt}
	\caption{\small
		Examples from \emph{CIM}.
		Here are some instances of \emph{Kate Winslet} and \emph{Leonardo DiCaprio}
		from different movies in \emph{CIM}.
	}
	\label{fig:dataset_jack_rose_hard}
\end{figure}

\begin{table}[!thb]
	\centering
	\begin{tabular}{|c|c|c|}
		\hline
		Dataset                                                               & PIPA\cite{zhang2015beyond}   & CIM (ours) \\
		\hline\hline
		Images                                                                & 37,107 & 72,875     \\
		\hline
		Indentities                                                           & 2,356  & 1,218      \\
		\hline
		Instances                                                             & 63,188 & 150,522    \\
		\hline
		\begin{tabular}[c]{@{}c@{}}Instances\\ (except "others")\end{tabular} & 51,751 & 77,598        \\
		\hline
		Avg/identity                                                          & 26.82  & 63.70      \\
		\hline
	\end{tabular}
	\vspace{-5pt}
	\caption{\small Statistics of \emph{CIM} compared with \emph{PIPA}.}
	\label{tab:dataset_stats}
\end{table}

In addition to photo albums, the proposed method can also be applied
to other settings with strong contexts, \eg~recognizing actors in movies.
To test our method in such settings, we constructed
the \emph{Cast In Movies (CIM)} dataset from $192$ movies.
We divide each movie into shots using an existing technique~\cite{apostolidis2014fast},
sample one frame from each shot, and retain all those that contain persons.
This procedure results in a dataset with $72,875$ photos.

We manually annotated all person instances in these photos with bounding boxes
for the body locations.
We also annotated the identities of those instances that correspond to
the $1218$ main actors\footnote{The main actors are chosen according
to two criteria: 1) ranked top 10 in the cast list of IMDb for the
corresponding movie, and 2) occur for more than 5 times in our sampled frames.}.
In this way, we obtained $77,598$ instances
with known identities, while other instances are labeled as ``others''.
Figure \ref{fig:dataset_jack_rose_hard} shows some examples of our dataset.
Table~\ref{tab:dataset_stats} shows the statistics of \emph{CIM}
in comparison with \emph{PIPA}~\cite{zhang2015beyond}.
To our best knowledge, \emph{CIM} is the first large-scale dataset for person
recognition in movies.



\section{Experiments}
\label{sec:experiment}

We tested our method on both
\emph{PIPA}~\cite{zhang2015beyond}, a dataset widely used for person recognition,
and \emph{CIM}, our new dataset presented above.

\subsection{Experiment Setup}
\label{subsec:exp_setup}

\begin{table*}[t]
	\centering
	\vspace{-10pt}
	\begin{tabular}{c|cccc|cccc}
		\hline
		\multirow{2}{*}{Split} & \multicolumn{4}{|c|}{Existing Methods on PIPA} & \multicolumn{4}{|c}{Ours} \\\cline{2-9}
		& PIPER~\cite{zhang2015beyond} & Naeil~\cite{joon2015person}  & RNN~\cite{li2016sequential} & MLC~\cite{li2016multi} & Baseline & +RANet & +RANet+P & +RANet+P+E \\\hline
		original & 83.05\% & 86.78\% & 84.93\% & 88.20\% & 82.79\% & 87.33\% & 88.06\% & \textbf{89.73\%}  \\\hline
		album    & -       & 78.72\% & 78.25\% & 83.02\% & 75.24\% & 82.59\% & 83.21\% & \textbf{85.33\%}  \\\hline
		time     & -       & 69.29\% & 66.43\% & 77.04\% & 66.55\% & 76.52\% & 77.64\% & \textbf{80.42\%}  \\\hline
		day      & -       & 46.61\% & 43.73\% & 59.77\% & 47.09\% & 65.49\% & 65.91\% & \textbf{67.16\%}  \\\hline
	\end{tabular}
	\vspace{-5pt}
    \caption{\small Comparison of the accuracies of different methods on PIPA,
        under different splits of the query and gallery sets.}
	\vspace{-10pt}
	\label{tab:result_pipa}
\end{table*}

\paragraph{Evaluation protocols}
The \emph{PIPA} dataset is partitioned into three disjoint sets:
training, validation and test sets.
The test set is further split into two subsets,
one as the gallery set and the other as the query set.
To evaluate a method's performance, we first use it to predict the identities
of the instances in the query set and compute the prediction accuracy.
Then, we switch the gallery and the query set and compute the accuracy
in the same way. The average of both accuracies will be reported as the
performance metric.

There are four different ways to split the test set,
namely \emph{original}, \emph{album}, \emph{time}, and \emph{day},
for evaluating an algorithm under different application scenarios.
In the \emph{original} setup of \emph{PIPA}~\cite{zhang2015beyond},
a query instance may have similar instances in the gallery.
\cite{joon2015person} defines the other three splits, which are
more challenging.
For example, \emph{day} split requires that the query
and the gallery instances of the same subject need to have notable differences
in visual appearance.
For \emph{CIM}, we follow the rule in~\cite{zhang2015beyond},
dividing it into three disjoint subsets respectively for training,
validation, and testing. Also, the test set is randomly split into
a gallery set and a query set.

\vspace{-11pt}
\paragraph{Implementation Details}
We use four regions of each instance:
\emph{face}, \emph{head}, \emph{upperbody}, and \emph{body}.
\emph{PIPA} provides the head locations, while \emph{CIM} provides the
locations of whole body.
Other regions are obtained by simple geometric rules based on the
results from a face detector~\cite{zhang2016joint} and OpenPose~\cite{cao2016realtime}.
Note that we only keep those bounding boxes that lie mostly within the photo.
For those regions that are largely invisible, we simply use a black image
to represent their appearance.
We will see that our RANet can learn to assign such regions
with negligible weights in our experiments.
We adopt ResNet-50~\cite{he2016deep} as our base model
and train the feature extractor with OIM loss~\cite{xiao2017joint}.
We chose design parameters empirically on the validation set.
The coefficients $\alpha$ and $\beta$ in Eq.\eqref{eq:jointobj} are set to
$0.05$ and $0.01$. The number of events $K$ is set to $300$ for both \emph{PIPA} and \emph{CIM}.

\subsection{Results on PIPA}
\label{subsec:exp_pipa}

We set up a \emph{baseline} for comparison, which relies on a \emph{uniformly}
weighted combination of visual cues from all regions, where the weights
are optimized by grid search.
We tested three configurations of the proposed methods:
\textbf{(1) +RANet:} This config combines region-specific scores following
Eq.\eqref{eq:vmatch}, using the instance-dependent weights from
the Region Attention Network (see Sec.~\ref{sec:vmatch}).
\textbf{(2) +RANet+P:} In addition to the visual matching score RANet,
it also uses the person-person relations in joint inference.
\textbf{(3) +RANet+P+E: } This is the full configuration of our framework,
which takes visual matching, person-person relations, and person-event relations
into account.
Moreover, we also compared with four previous methods:
PIPER~\cite{zhang2015beyond},
Naeil~\cite{joon2015person},
RNN~\cite{li2016sequential}, and
MLC~\cite{li2016multi}.

Table~\ref{tab:result_pipa} shows the results under all the four splits,
from which we can see that:
(1) RANet, with adaptive weights, significantly outperforms the baseline with
uniform weights. On the most challenging \emph{day split}, it remarkably
raises the accuracy from $47.09\%$ to $65.49\%$.
(2) With our proposed joint inference method, the use of social contexts
leads to consistent improvement across all splits.
(3) Our method also outperforms all previous works,
including the state-of-the-art MLC~\cite{li2016multi}, by a considerable
margin on all splits.
Particularly, the performance gain is especially remarkable on the most
challenging \emph{day split} ($67.16\%$ with ours vs. $59.77\%$ with MLC).

\begin{figure}
	\centering
	\includegraphics[width=\linewidth]{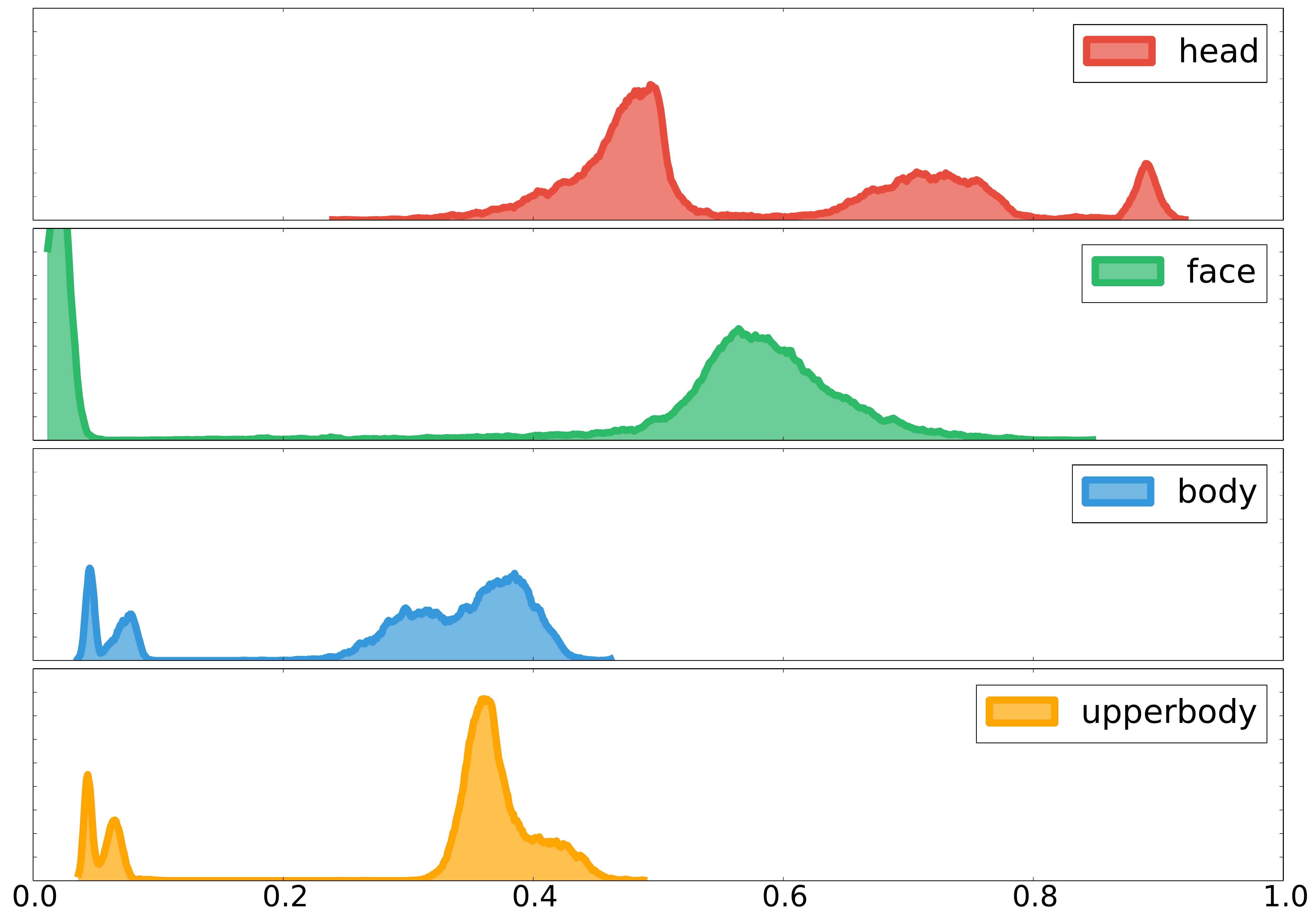}
	\vspace{-15pt}
	\caption{\small
		Weight distributions of different regions on \emph{PIPA}.
	}
	\label{fig:weight_distribution}
	\vspace{-5pt}
\end{figure}

\begin{figure}
	\centering
	\includegraphics[width=\linewidth]{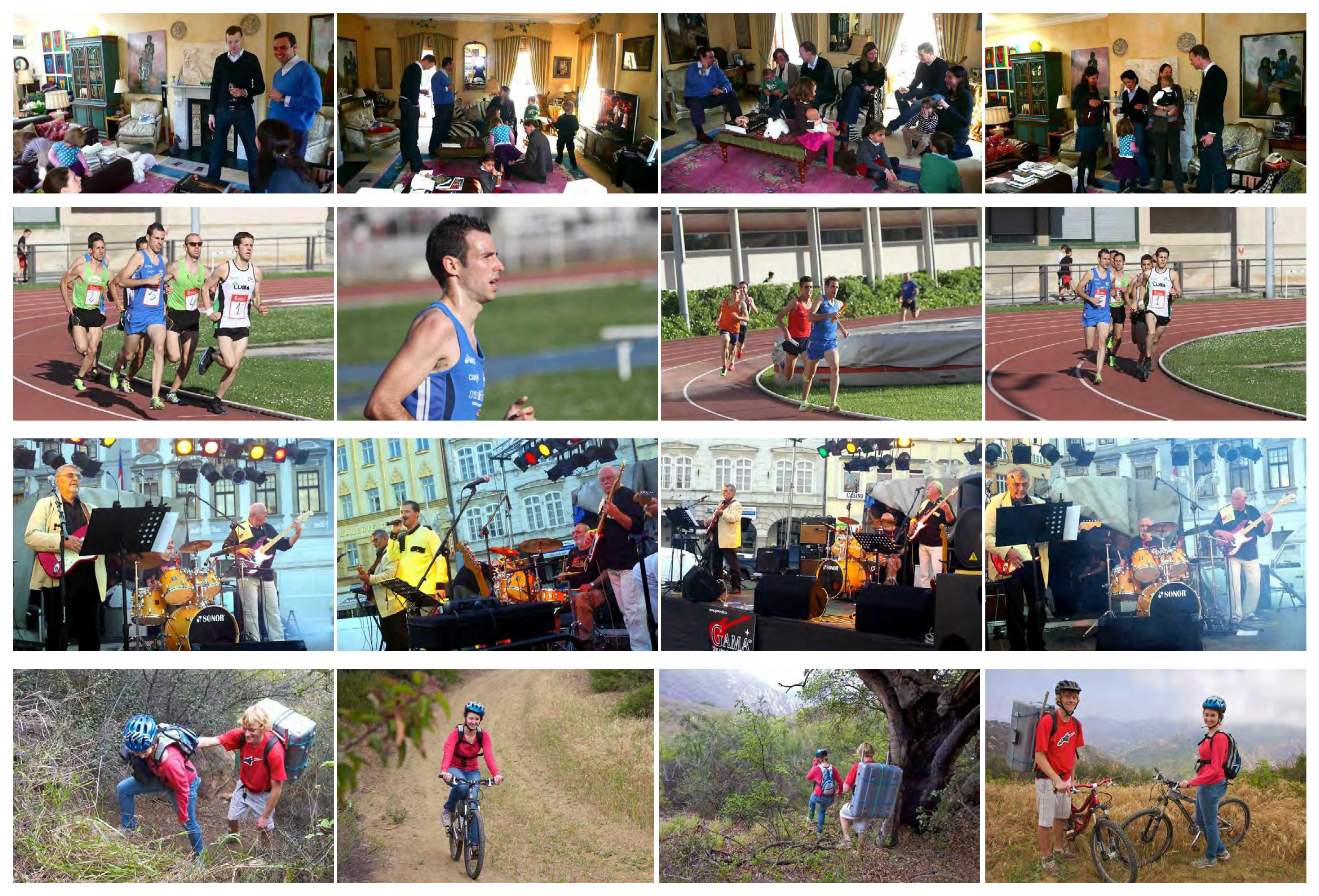}
	\vspace{-15pt}
	\caption{\small
		Example events automatically discovered from \emph{PIPA} by joint inference.
		Images in each row belong to the same ``event''.
	}
	\label{fig:event_sample}
	\vspace{-10pt}
\end{figure}

\begin{figure}[t]
	\centering
	\includegraphics[width=\linewidth]{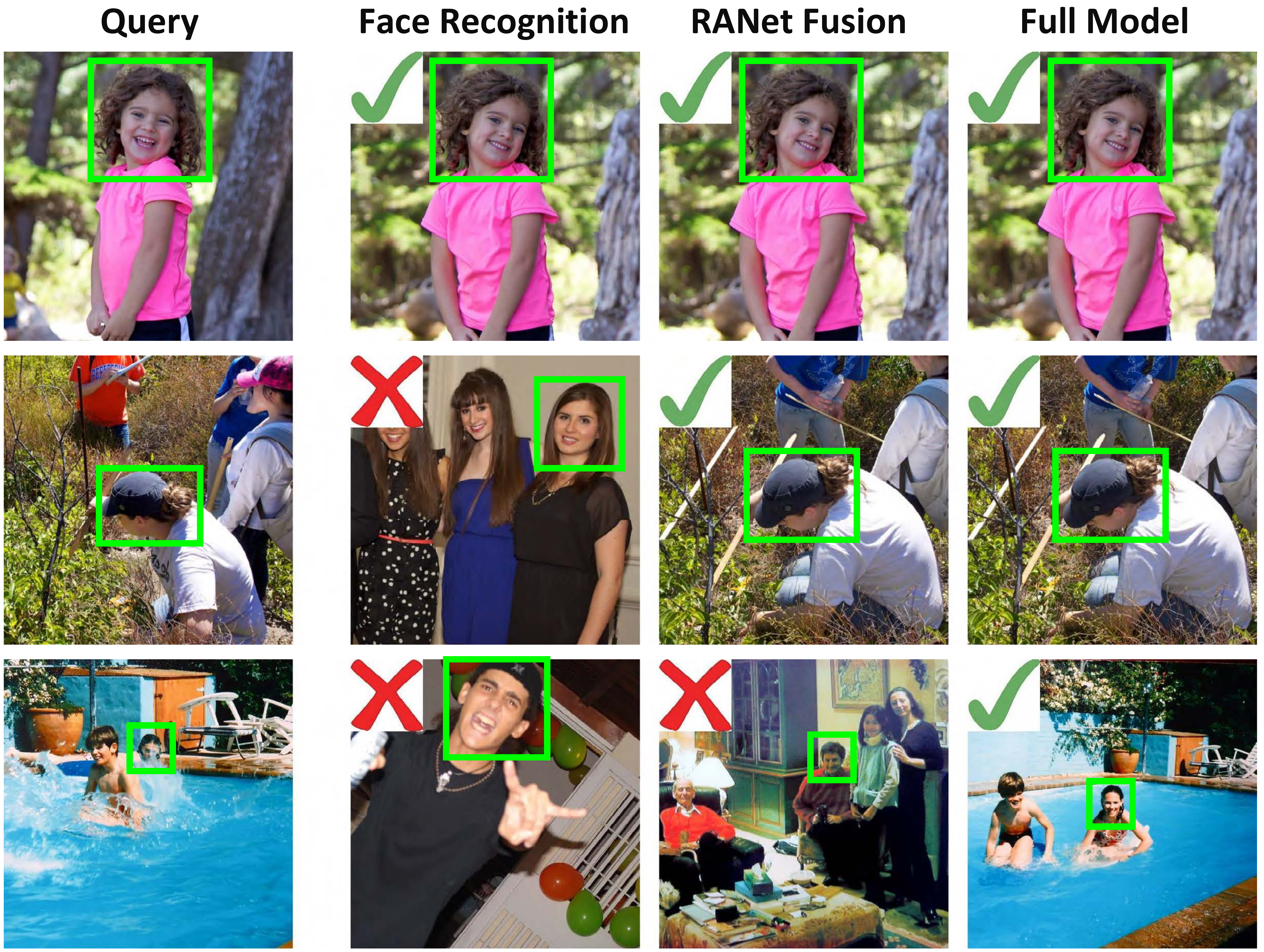}
	\caption{\small
		Example photos and recognition results.
		For each photo,
		the mark at the top left corner indicates whether
		the corresponding method predicts correctly
		for the highlighted instance.
	}
	\label{fig:case_study}
	\vspace{-10pt}
\end{figure}

\vspace{-9pt}
\paragraph{Analysis on RANet}

Table~\ref{tab:result_pipa} clearly shows the effectiveness of the
Region Attention Network (RANet).
To learn more about the RANet,
we study the distributions of region-specific weights on the test set of \emph{PIPA},
and show them in Figure~\ref{fig:weight_distribution}.
This study reveals some interesting characteristics of RANet:
(1) For each of the following region types,
\emph{face}, \emph{body}, and \emph{upper body}, there exist a fraction
of instances with very low weights because the particular regions of them are out of scope.
(2) The average weight of \emph{faces} is the highest among all region types.
This is not surprising, as \emph{faces} are often the strongest indicators
of identities when they are visible.
(3) A small portion of instances have very high weights assigned to
the \emph{head} regions, because for such instances all other parts
are largely invisible.

\vspace{-9pt}
\paragraph{Analysis on Event}
Events are automatically discovered during joint inference and they
play an important role in person recognition.
Figure~\ref{fig:event_sample} shows some example events with
their associated photos.
We can see that our method can discover events in a reasonable way,
and they can provide strong prior in a considerable portion of cases.
More examples will be provided in the supplemental materials.

\vspace{-9pt}
\paragraph{Case Study}
Figure~\ref{fig:case_study} shows some photos and associated recognition
results.
We can see that
1) For an instance with frontal and clear face (1st row),
all methods predict correctly.
2) When the face is not clearly visible (2nd row),
our method with RANet can still correctly recognize the person with
other visual cues, \eg~the head or upper body.
3) For the most challenging case where all visual cues fail (3rd row),
our full model can still make a correct prediction by exploiting
the social context.

\subsection{Results on CIM}
\label{subsec:exp_cim}

Table~\ref{tab:result_cim} shows the results on \emph{CIM},
which again demonstrates the effectiveness of our approach.
Only with RANet, it already outperforms the baseline (with uniform weighting)
by over $4\%$. The whole framework, with social context taken into account,
further improves the accuracy ($6.3\%$ higher than the baseline).
Recognition results on example photos will be provided in the supplemental
materials.
It is also worth noting that the accuracies we obtained on \emph{CIM}
are generally lower than those on \emph{PIPA}, implying that this is a
more challenging dataset which can help to drive the progress on this task.

\begin{table}[t]
	\centering
	\begin{tabular}{c|c|c|c}
		\hline
		\small{Baseline} & \small{+RANet} & \small{+RANet+P} & \small{+RANet+P+E} \\\hline
		68.12\% & 71.93\% & 72.56\% & \textbf{74.40\%} \\\hline
	\end{tabular}
	\vspace{-5pt}
	\caption{\small Performance on \emph{CIM}}
	\label{tab:result_cim}
	\vspace{-10pt}
\end{table}

\subsection{Computational Cost Analysis}
\label{subsec:exp_cost}

Our method obtains
the improvement on recognition accuracy with substantially
lower computing cost compared to some previous works.
Note that PIPER~\cite{zhang2015beyond} uses more than 100 deep CNNs
and Naeil~\cite{joon2015person} uses 17 deep CNNs for feature extraction.
While our model uses only 4 CNNs and a fusion module whose computing cost
is negligible.
Although MLC~\cite{li2016multi} uses just 3 deep CNNs for feature extraction,
it additionally requires to train thousands of group-specific SVMs, which
is also a costly procedure.

Compared with the CNN-based feature extraction components, the cost of the joint
estimation and inference procedure is insignificant. Particularly, it takes
about \emph{30 minutes} to perform inference over the whole test set of \emph{PIPA},
with one single $2.2$ GHz CPU, while the feature extractors take over $40$ hours
to detect regions and compute CNN features for all test photos,
with a Titan X GPU.


\section{Conclusions}
\label{sec:conclusion}

We presented a new framework for person recognition, which integrates
a Region Attention Network to adaptively combine visual cues and
a model that unifies person identification and context learning
in joint inference.
We conducted experiments on both PIPA and a new dataset CIM constructed
from movies. On PIPA, our method consistently outperformed previous
state-of-the-art methods by a notable margin, under all splits.
On CIM, the new components developed in this work also demonstrated
strong effectiveness in raising the recognition accuracy.
Both quantitative and qualitative studies showed that adaptive combination
of visual cues is important in a generic context and that the social context
often conveys useful information especially when the visual appearance
causes ambiguities.


\section{Acknowledgement}
\label{sec:acknow}

This work is partially supported by the Big Data Collaboration Research grant from SenseTime Group (CUHK Agreement No. TS1610626), the General Research Fund (GRF) of Hong Kong (No. 14236516).
We are grateful to Shuang Li and Hongsheng Li for helpful discussions.

{\small
\bibliographystyle{ieee}
\bibliography{main}
}

\end{document}